\documentclass[conference]{IEEEtran}

\pdfoutput=1

\usepackage[pdftex]{graphicx}
\usepackage{url}
\usepackage{amsmath}
\usepackage{amssymb}
\usepackage{color}

\usepackage{epstopdf}
\usepackage{tabularx}
\graphicspath{{./images/}}
\usepackage{hyperref}
\usepackage{subfigure}
\usepackage{gensymb}
\usepackage{lipsum}
\usepackage[export]{adjustbox}
\usepackage{flushend}
\usepackage{caption}
\newcommand{\minisection}[1]{\vspace{0.0in} \noindent {\bf #1}\ \ }

\begin{document}

\title{Sparse Radial Sampling LBP for Writer Identification}
\author{\IEEEauthorblockN{Anguelos Nicolaou\IEEEauthorrefmark{1},
Andrew D. Bagdanov\IEEEauthorrefmark{1},
Marcus Liwicki\IEEEauthorrefmark{2}, and
Dimosthenis Karatzas\IEEEauthorrefmark{1}}
\vspace{-.35cm}
\IEEEauthorblockA{\\\IEEEauthorrefmark{1}Computer Vision Center, Edifici O, Universitad Autonoma de Barcelona,Bellaterra, Spain
\\
\IEEEauthorrefmark{2}DIVA research group, Department of Informatics, University of Fribourg, Switzerland \\
Email: \href{mailto:anguelos@cvc.uab.es}{anguelos@cvc.uab.es},
\href{mailto:bagdanov@cvc.uab.es}{bagdanov@cvc.uab.es}, 
\href{mailto:marcus.liwicki@unifr.ch}{marcus.liwicki@unifr.ch},
\href{mailto:dimos@cvc.uab.es}{dimos@cvc.uab.es}
}

}

\maketitle

\begin{abstract}
In this paper we present the use of Sparse Radial Sampling Local Binary Patterns, a variant of Local Binary Patterns (LBP) for text-as-texture classification.
By adapting and extending the standard LBP operator to the particularities of text we get a generic text-as-texture classification scheme and apply it to writer identification.
In experiments on CVL and ICDAR 2013 datasets, the proposed feature-set demonstrates 
State-Of-the-Art (SOA) performance.
Among the SOA, the proposed method is the only one that is based on dense extraction of a single local feature descriptor. This makes it fast and applicable at the earliest stages in a DIA pipeline without the need for segmentation, binarization, or extraction of multiple features.

\end{abstract}

\IEEEpeerreviewmaketitle

\section{Introduction}

Writer identification is the problem of identifying the authorship of
text samples based on an index of examples of text written by
known authors~\cite{Bulacu:2007}. It has a long tradition in forensics
where it has been accepted by the court as evidence for more than a century.
From a pattern recognition perspective, three variations of the problem of writer identity are defined: writer identification, writer verification, and writer retrieval.
Writer identification is the most popular of these, and in most cases a method can be modified from solving one to solving another with little effort.

From a Document Image Analysis (DIA) perspective there are other
applications such as scribe identification for historical
documents. Handwriting has been considered as a
behavioral biometric~\cite{delta2014,bertolini2013}, although
experiments on disguised handwriting have proved to confuse both
graphonomists, and automatic systems~\cite{malik2013}. Despite the attention writer
identification has received from DIA researchers in recent years, it
remains a difficult problem due to variations in writing style and
conditions, and the myriad problems arising from variations in image
quality due to document degradation and other incidental factors.

LBP are dense local texture descriptors that can be used to describe the local structure of images~\cite{ojala2002lbp}.
They have been successfully applied to many of the major computer vision problems, as well also been applied to specific problems in
Document Image Analysis, including optical font recognition and writer identification.
LBP were originally designed for graylevel images, and despite their widespread application to bilevel document images, it remains unclear how LBP should be computed on such images in order to remain discriminative and robust to noise.

In this paper we introduce Sparse Radial Sampling LBP (SRS-LBP), a variant of LBP that is better suited for the
task of writer identification and text-as-oriented-texture
classification in bilevel images in general. Our main contribution is
the introduction of sparse radial sampling of the circular patterns
used for LBP construction.  This allows sampling of patterns up to
very large radii for each pixel at low computational cost, and we can
also avoid vocabulary compression techniques such as rotation
invariant or uniform patterns commonly applied to standard LBP
representations.  We show that using a single local descriptor, our
SRS-LBP variant densely extracted and pooled over the entire image,
results in a low-dimensional feature representation that yields
SOA performance at a fraction of the cost of other
techniques. Our representation is compact and extremely efficient to
compute.

In the next section we describe work from the literature related to our application of LBP to writer identification.
We describe the modified LBP we use and how we apply it to writer identification in Section~\ref{sec:writer_identification}.
In Section~\ref{sec:experiments} we report on experiments we performed comparing our approach with the SOA on two standard writer identification benchmarks, and in section~\ref{sec:conclusion} we conclude with a discussion of our contribution and ongoing work.

\section{Related work}
\label{sec:related_work}

\begin{figure*}
\centering
\setlength\tabcolsep{.25cm}
\begin{tabular}{cccc}
\includegraphics[width=.2\textwidth]{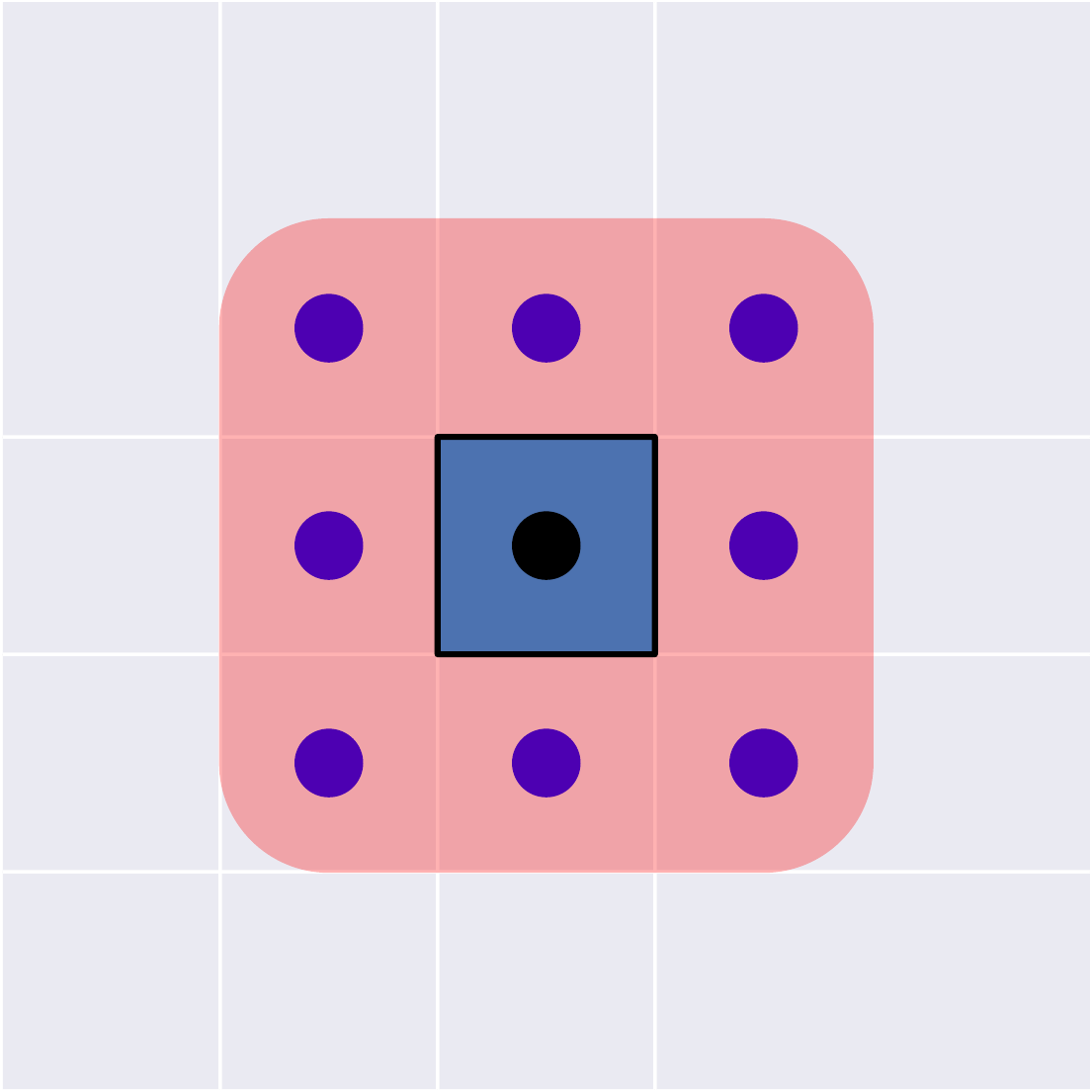}&
\includegraphics[width=.2\textwidth]{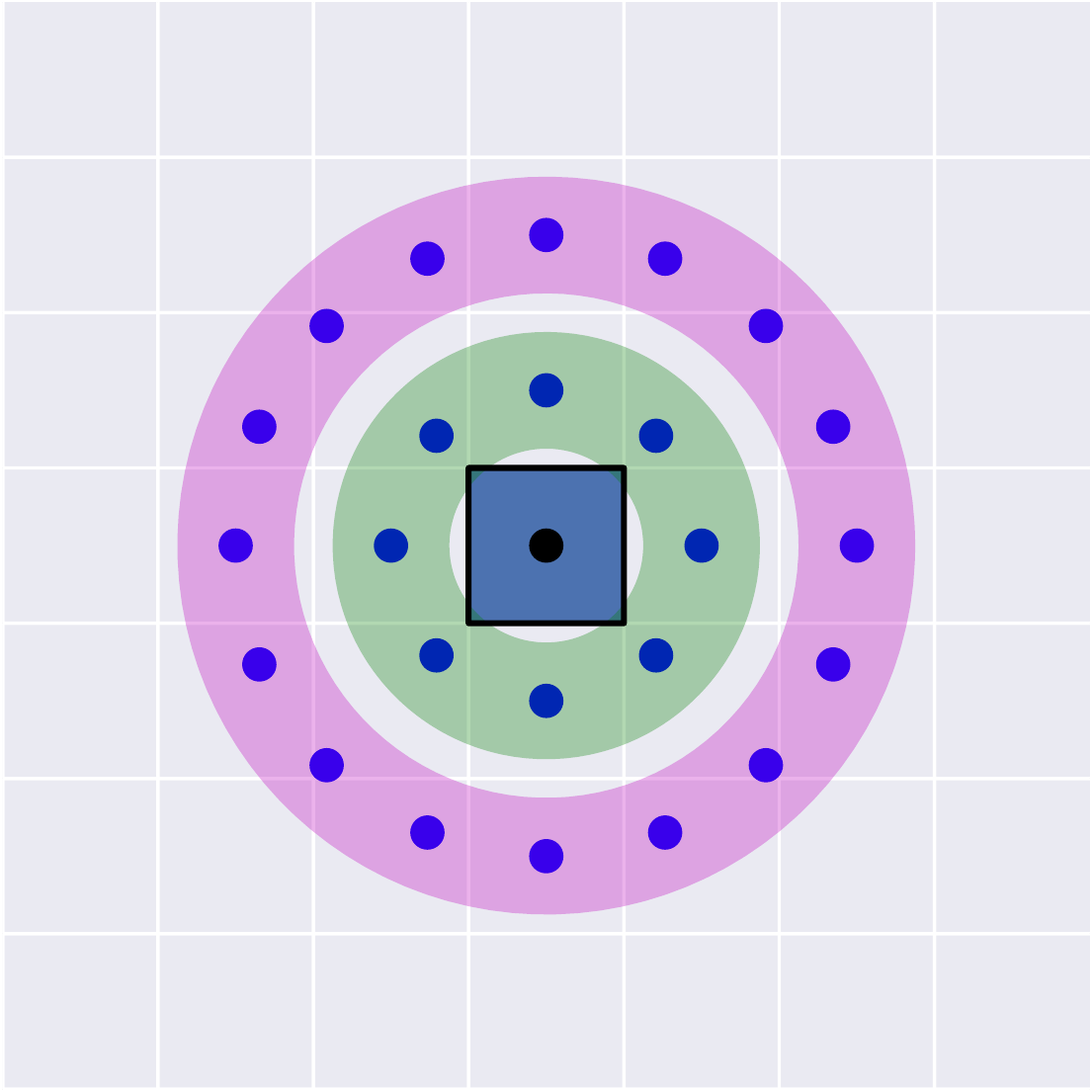} &
\includegraphics[width=.2\textwidth]{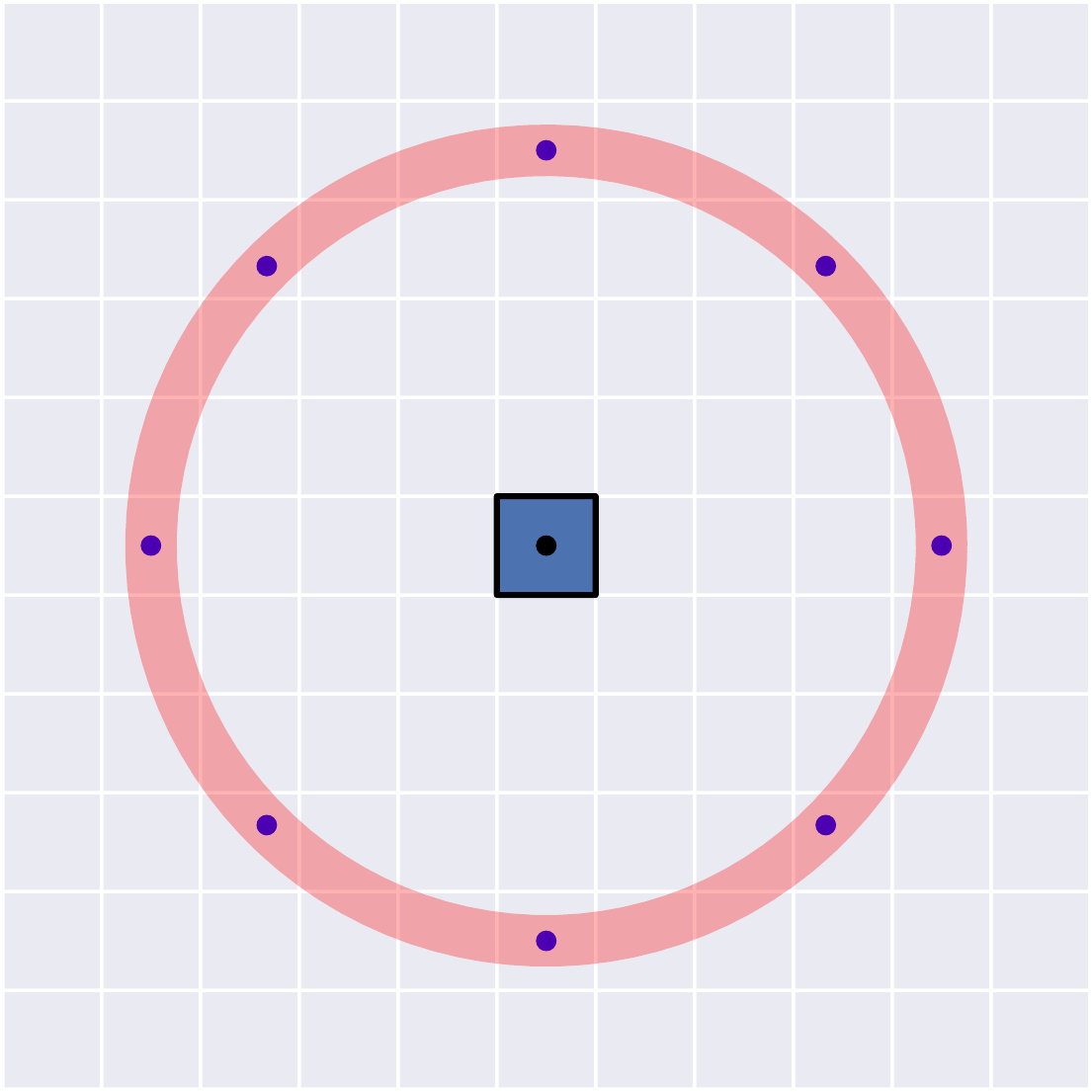} &
\includegraphics[width=.2\textwidth]{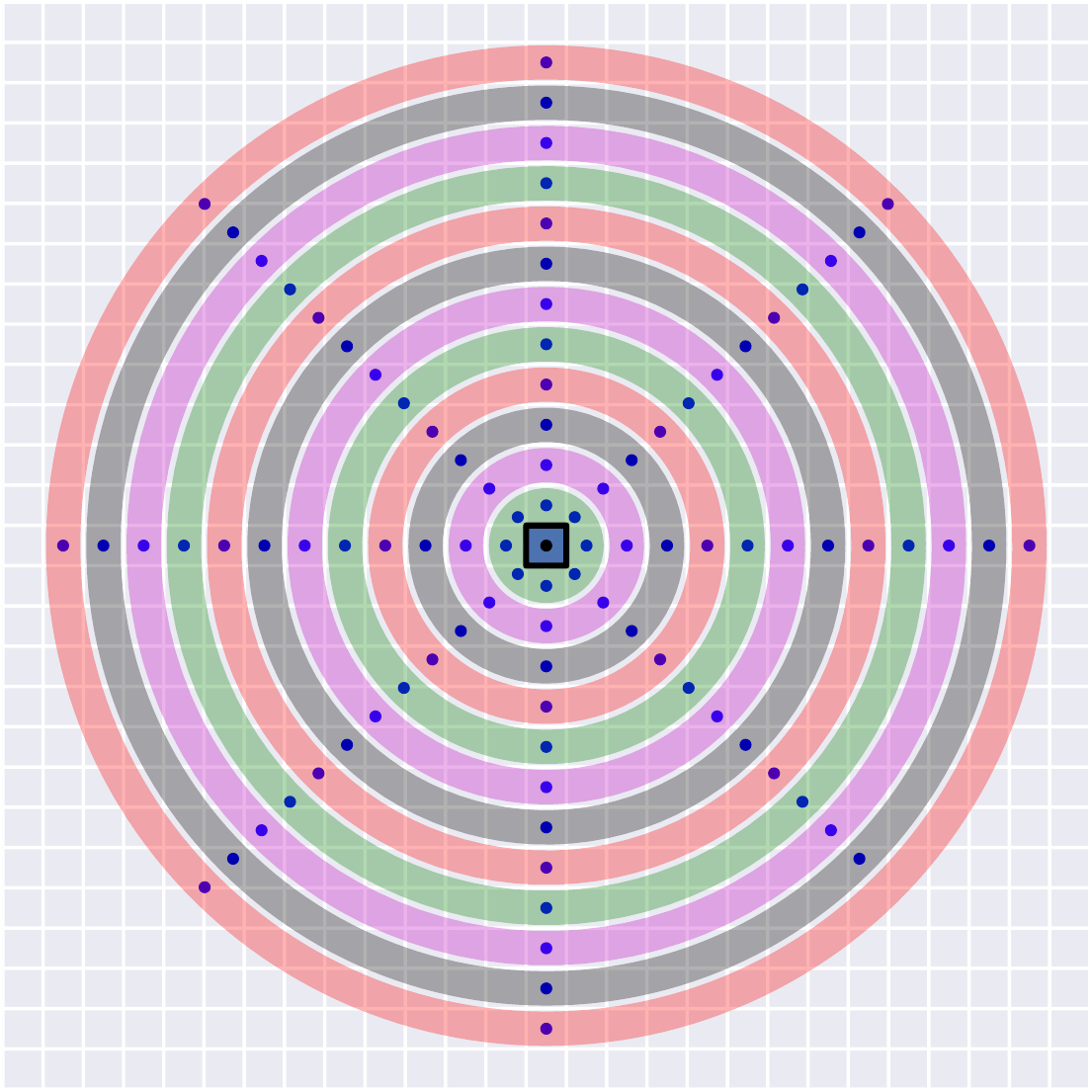} \\
(a) Original $\mbox{LBP}_{3 \times 3}$  & 
(b) $ \left[\mbox{LBP}_{1,8}, \mbox{LBP}_{2,16} \right]$ & 
(c) $ \mbox{LBP}_{4,8}  $ &
(d) $ \left[\mbox{LBP}_{1,8},..., \mbox{LBP}_{12,8} \right]$ 
\end{tabular}
\vspace{-.2cm}
\caption{Baseline (a),(b) and Proposed (c),(d) LBP transform sampling patterns. (b) and (d) mark several LBP used in in conjunction.}
\label{fig:patterns}
\end{figure*}

\subsection{Local Binary Patterns}
LBP characterize local image patches using binary codes that encode the relationship between a central pixel and its neighbors~\cite{ojala2002lbp}.
LBP feature extraction usually consists of computing LBP descriptors at each pixel of an image to create an image of integer valued codes, followed by pooling of these codes into a
histogram~\cite{ojala2002lbp}.
LBP have been successfully applied to many of the major computer vision problems such as face recognition~\cite{ahonen2006}, and human detection~\cite{mu2008discriminative}.

In DIA, LBP have been used for text detection of text in television streams~\cite{anthimopoulos2008}, for printed script detection~\cite{ferrer2013}, and in~\cite{wei2014} LBP were compared to other features in a feature selection process for historical document layout analysis and were selected as best across all datasets.
The authors used an earlier proponent of the method presented in this paper for Arabic font recognition~\cite{nicolaou2014}. 

\subsection{Writer Identification}
Automatic writer identification has been researched for many decades.
In 2007 Schomaker et al. propose allographic features for writer
identification, and for an overview of previous work on writer
identification, we refer to their excellent
survey~\cite{Bulacu:2007}. Chawki et al. applied run length features in 2010 for Arabic writer identification~\cite{chawki2010}.
Jain et al. used clusters of pseudo-letters~\cite{jain2013} and fusion of features~\cite{jain2014}.
LBP have also been used for writer identification: Du et al. extracted LBP features from the wavelet transform of Chinese hand-writing samples~\cite{du2010wavelet}, and Bertolini et al.~\cite{bertolini2013} used LBP in a comparative study with Local Phase Quantization and concluded that LPQ performed better than their LBP variant.

\subsection{Our contribution with respect to the State-Of-The-Art}
\label{sec:contribution}

Our adaptation of the LBP consists of replacing the sign operator with a threshold statistically derived from each image, the use of sparse radial sampling at each radius, and the use of very large radii when computing LBP.
Though we concatenate LBP histograms extracted at many radii, sparse radial sampling ensures that the final LBP features are compact.

We apply our SRS-LBP to writer identification using a standard LBP pipeline.
SRS-LBP are computed at each location in an image, and these features are pooled over the entire page image. 
Our approach requires no character segmentation and is based on a single, compact feature that is extremely efficient to
extract. 
In this sense it stands out with respect to SOA approaches based on complex character segmentation, clustering, and extraction of multiple feature descriptors~\cite{jain2014}.

\section{Writer identification with LBP}
\label{sec:writer_identification}

In this section we describe our approach to LBP extraction based on sparse radial sampling. We follow the development and notation of~\cite{ojala2002lbp}.

\subsection{The LBP transform}

LBP feature extraction consists of two principal steps: the LBP transform, and the pooling of LBP into a histogram representation of an image.
The LBP transform maps each pixel to an integer code representing the relationship between the center pixel and the pixels of its neighbourhood.
It encapsulates the local geometry at each pixel by encoding binarized differences with pixels of its local neighbourhood:
\begin{equation}
\textrm{LBP}_{P,R,t}=\sum_{p=0}^{P-1}s_t(g_p-g_c)*2^{p},
\label{eq:lbpsum}
\end{equation}
where $g_c$ is the central pixel being encoded, $g_p$ are $P$ symmetrically and uniformly sampled points on the periphery of a circular area of radius $R$ around $g_c$, and $s_t$ is a binarization function parametrized by $t$. 
The sampling of $g_p$ is performed with bilinear interpolation.
The use of local differences in~(\ref{eq:lbpsum}) endows LBP with a degree of illumination invariance.

In our LBP definition, $s$ is a simple threshold:
\begin{equation}
\begin{split}
s_t(x) = \left\{
     \begin{array}{lr}
       1 & : x \geq t \\
       0 & : x < t
     \end{array}
   \right.,
\end{split}
\label{eq:lbpcmp}
\end{equation}
where $t$, which in the standard definition is considered zero, is a parameter that determines when local differences are considered ``big enough'' for consideration.

$\textrm{LBP}_{P,R,t}$ can be seen as a transform from the graylevel
domain to a domain of discrete labels encoded over a vocabulary of
$2^P$ integers.
In Fig.~\ref{fig:patterns} sampling patterns of popular and proposed LBP can be seen where $g_c$ is marked as a black dot and $g_p$ are marked as blue dots. 

\subsection{Sparse sampling LBP on bilevel images}
The original LBP was designed for graylevel images.
Though text images are often fundamentally bilevel by nature, the bilinear interpolation used to extract neighbouring pixel values $g_p$ renders pixels non-binary and standard LBP is (at least mathematically) applicable.
However, images of a bilevel nature such as text even when they are acquired on the graylevel domain do not benefit much from illumination invariance.
Another problem is that large $g_c-g_p$ differences are more rare than small ones and so treating both of them the same introduces noise.

Rather than use arbitrary or empirically derived threshold $t$ to re-binarize differences in the computation of LBP in~(\ref{eq:lbpsum}), we propose to apply Otsu's method to estimate optimal threshold $\hat{t}$ from the statistics of image differences themselves:
\begin{eqnarray}
\hat{t} = \arg \min \omega(d_{1,t})\sigma^2(d_{1,t})+\omega(d_{t,P})\sigma^2(d_{t,P})
\end{eqnarray}
where $d_{1,t}$ is the set of $|g_c-g_p|$ less than threshold $\hat{t}$,  $d_{t,P}$ is the set of $|g_c-g_p|$ greater than the threshold, $\omega$ is the probability of $d_{\dots}$ and $\sigma^2$ its variance.
The use of $\hat{t}$ yields a unified solution to both
of these problems. The Otsu threshold of the differences effectively
separates the significant differences from insignificant ones. Note
that this formulation works for bilevel and graylevel source
imagery.

Sparse radial sampling is integrated into our descriptor by holding
constant the number of points sampled at each radius:
\begin{eqnarray}
\textrm{SRS-LBP}_R = \textrm{LBP}_{8,R,\hat{t}}.
\label{eq:srslbp}
\end{eqnarray}
Keeping $P=8$ constant (i.e. sparse radial sampling) allows us to
sample more radii while maintaining a compact code.

\subsection{Processing pipeline}
\label{sec:histogram}
Our complete processing pipeline is comprised of the following steps:
\begin{enumerate}
\item \textbf{SRS-LBP transformation:} each image pixel is transformed
  to several SRS-LBP according to~(\ref{eq:lbpsum})
  and~(\ref{eq:srslbp}). This encodes the input image as several
  8-bit images (one for each radius).
\item \textbf{SRS-LBP pooling:} a histogram of SRS-LBP codes
  is computed for each radius. We discard the zero pattern
  which corresponds to foreground- and background-only patterns, and
  then L1 normalize and concatenate all histograms. The result is a
  block-normalized descriptor of size $256 \times |R|$, where $|R|$ is
  the number of radii.
\item \textbf{PCA projection}: the block-normalized descriptor is
  projected onto the first $N$ principal components computed through
  Principal Component Analysis (PCA).
\item \textbf{Normalization:} the Hellinger kernel is applied to the
  projected descriptor, followed L2 normalization. This combination
  has been shown to improve performance of a variety of image
  recognition techniques and specifically on writer
  identification~\cite{supervector2014}.
\end{enumerate}

Note that we use none of the standard vocabulary compression
techniques such as rotation invariance or uniformity used
in~\cite{ojala2002lbp} to make possible the usage of larger radii.
Avoiding these techniques is one of the main motivations for SRS-LBP
because, in the case of textual images specifically, there is
important information in the discarded patterns.

\section{Experimental results}
\label{sec:experiments}
\begin{figure*}
\centering
\begin{tabular}{ccc}
\includegraphics[width=.30\textwidth,height=.22\textwidth]{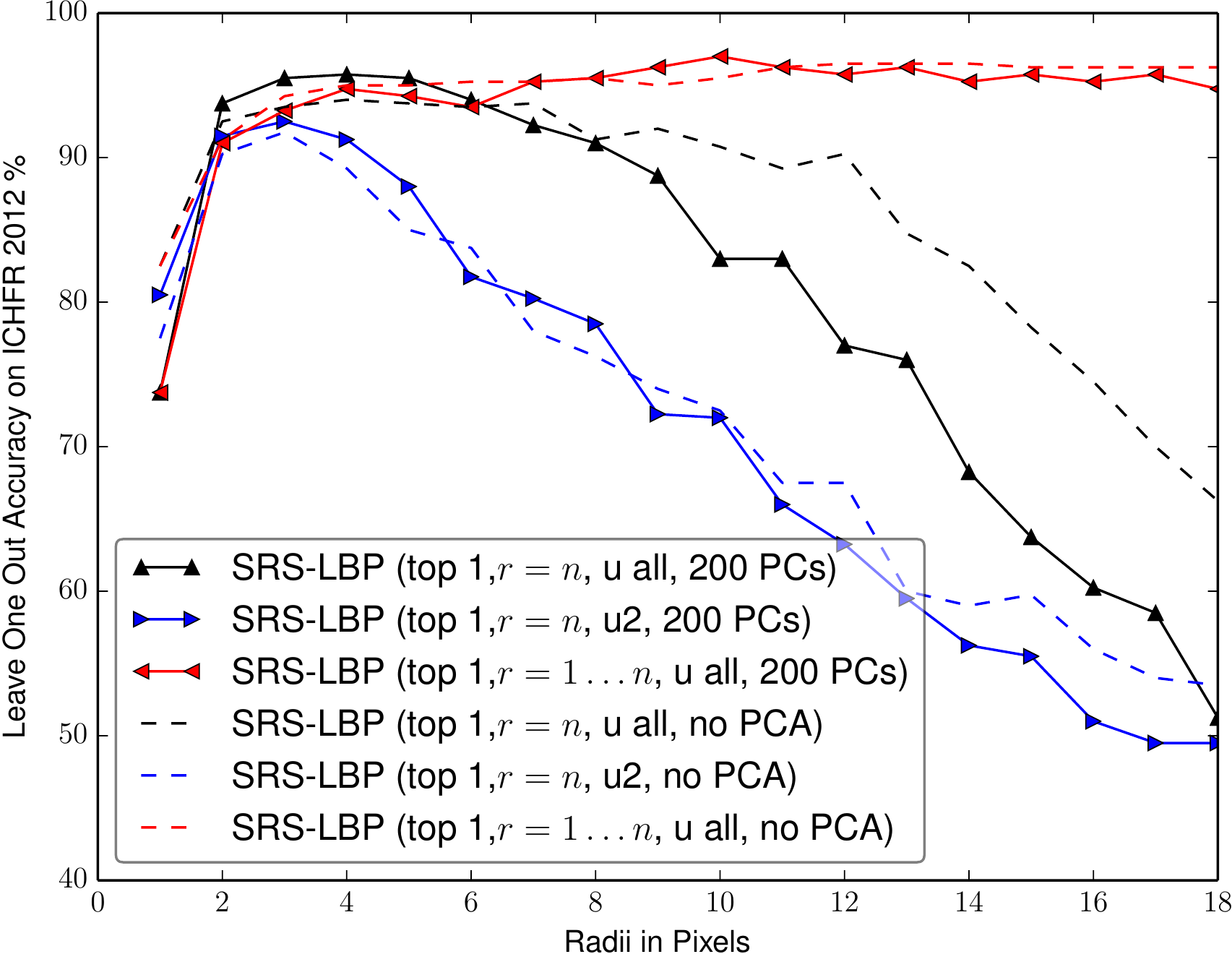} &
\includegraphics[width=.30\textwidth,height=.22\textwidth]{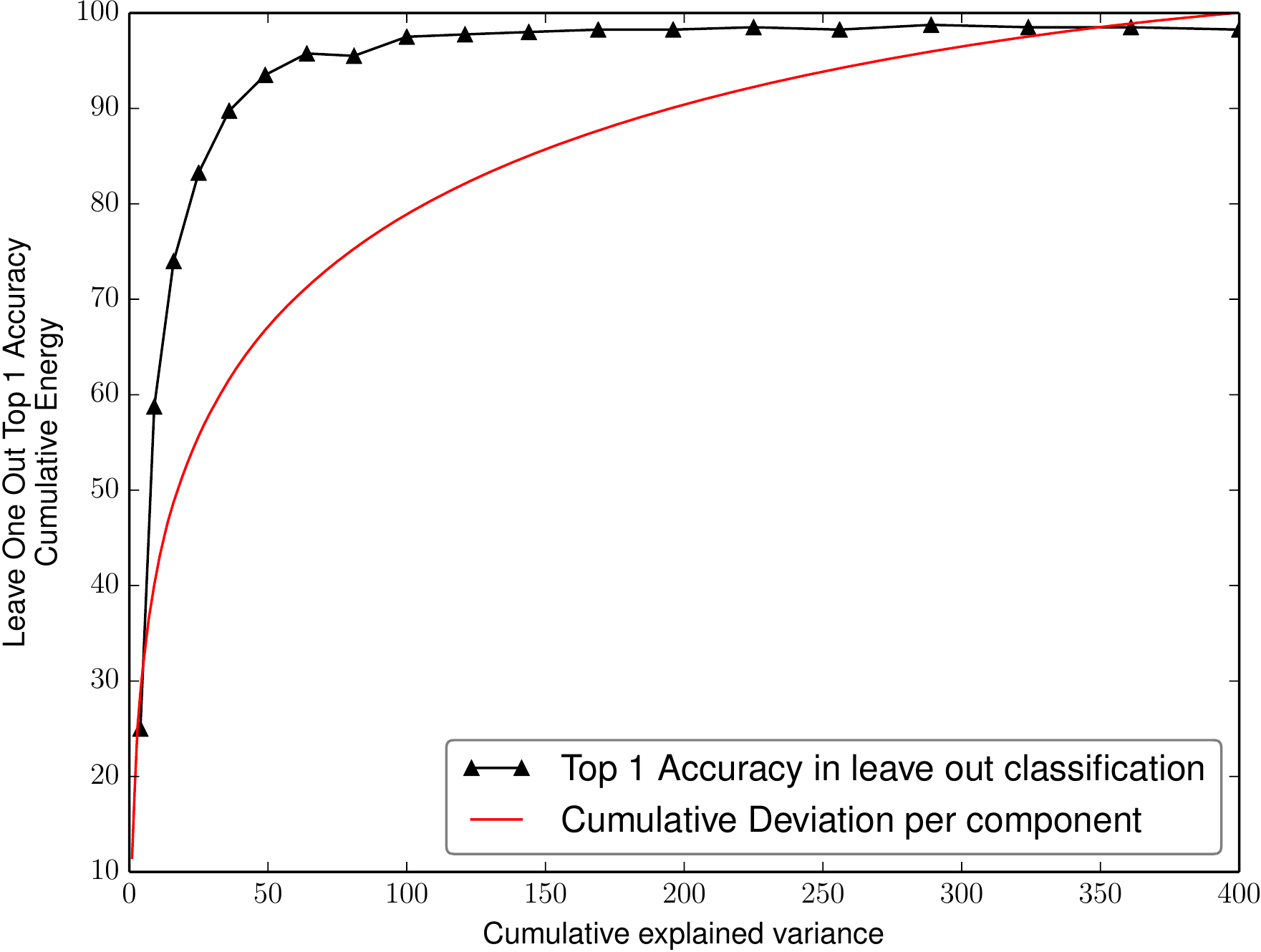} &
\includegraphics[width=.30\textwidth,height=.22\textwidth]{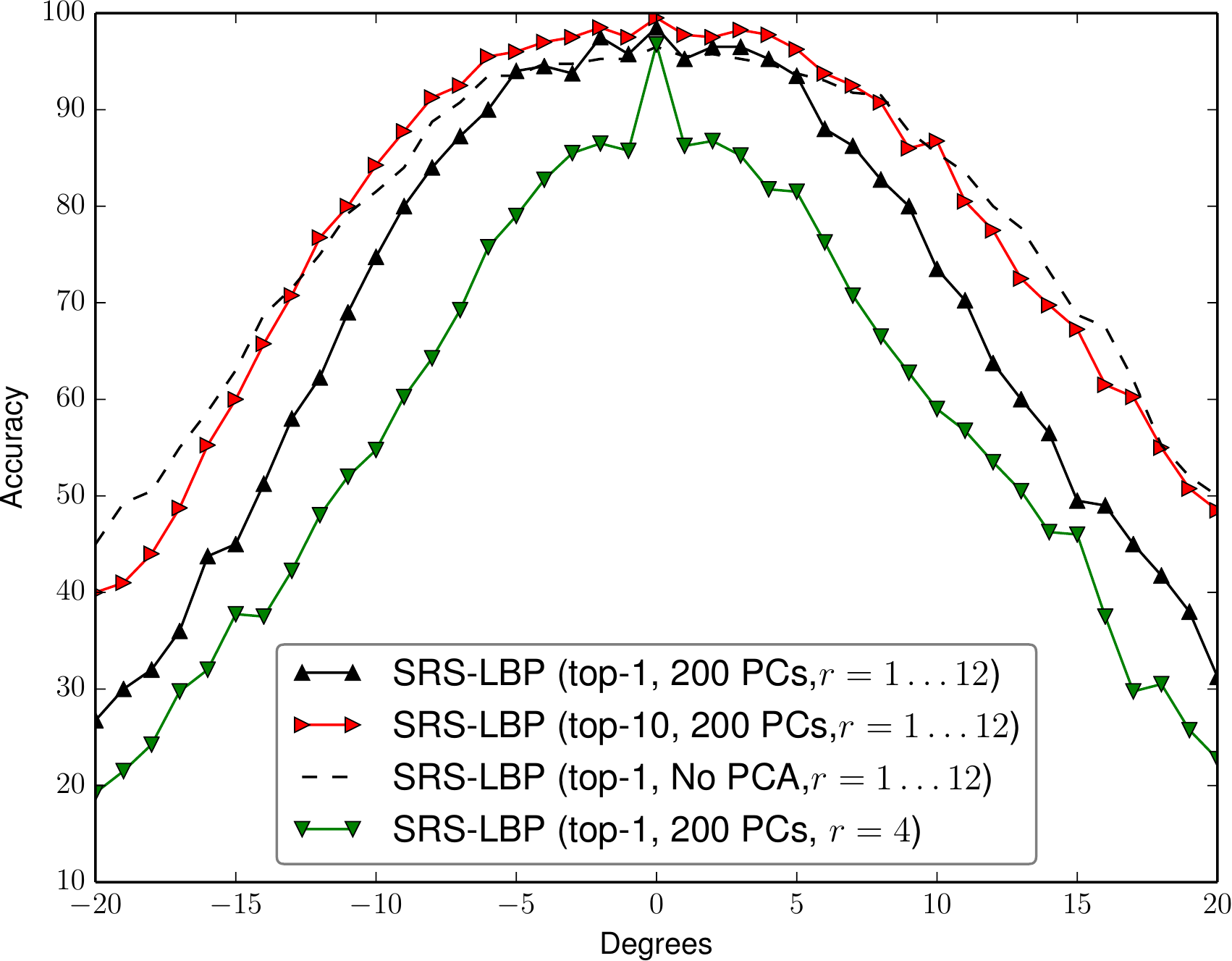} \\
(a) contribution of radii & (b) principal component contribution  & (c) rotation sensitivity \\
\end{tabular}
\vspace{-.2cm}
\caption{Parameter estimation and rotation sensitivity. (a) Individual
  and cumulative contribution of each radius to the top-1
  accuracy. (b) The cumulative contribution principal components have
  in top-1 accuracy. (c) The tolerance of SRS-LBP to rotations of the
  image.  }
\label{fig:qualitative}
\end{figure*}

In this section we report on a series of writer identification
experiments we performed to evaluate the potential of SRS-LBP and to compare its performance with the
SOA.

\subsection{Datasets}
We use a range of publicly available benchmark datasets for our
experimental evaluation.

\minisection{ICDAR 2013:} The dataset from the ICDAR 2013 competition
consists of 1,000 samples from 250 persons who each
contributed two samples in English and two in
Greek~\cite{icdar2013}. 

\minisection{CVL:} 
The CVL~\cite{cvl2013} dataset consists of 1,550 samples from 310
persons who contributed four samples in English and one in German. The
samples were acquired in color with different pens.  

\minisection{ICHFR 2012:} The ICHFR 2012 dataset consists of 400
samples from 100 subjects contributing two samples in English and two
samples in Greek~\cite{icfhr2012}. We use this dataset for
baseline performance analysis.

\subsection{Evaluation protocols}
Our approach uses a nearest neighbor classifier, and evaluation is based on leave-one-out cross validation. 
For each sample represented in feature space, we rank all remaining samples by their distance to that sample. Two important performance measures are the top-$n$ soft criterion, which means having any image of the same class as the query sample in the first $n$ most ranked results, and the top-$n$ \emph{hard} criterion which means having \emph{only} images of the same class as the query sample in first $n$ samples~\cite{icdar2013,icfhr2012}.

Comparison with the SOA is complicated by the wide
variety of evaluation protocols for writer identification used by
international benchmarks and contests.
Just indicatively on the three most recent competitions in writer identification, methods had to be, a similarity measurement~\cite{icfhr2012}, a trainable classifier~\cite{malik2013}, and a feature extraction method accompanied by a metric~\cite{icdar2013}.

In our evaluation we employ two evaluation protocols. 
One we refer to as \textbf{metric} and is a protocol totally compatible with measurements in~\cite{icfhr2012,icdar2013} (e.g. we only consider pairs of samples, never the entire dataset as a whole). 
The other we call \textbf{l1out} and is the average performance of a leave-one-out cross-validation in a trainable classifier sense which is compatible with~\cite{malik2013}. In practice the difference between \textbf{l1out} and \textbf{metric} is that \textbf{l1out} allows access to all the samples in the evaluation dataset while \textbf{metric} restricts access to each sample alone.

For our approach, the difference between the two amounts to whether PCA analysis was done on the evaluation dataset (and thus learning from it) or an independent dataset.
In all experiments other than comparison with the SOA, the evaluation protocol used is \textbf{l1out}. 
Since \textbf{metric} is a stricter protocol than \textbf{l1out}, all SOA performance numbers derived from a protocol that explicitly adheres to metric are marked with an asterisk (*).

\subsection{Baseline performance analysis}
\label{sec:experiments_baseline}

Here we report on a number of baseline experiments we performed to quantify the performance of our approach and to estimate key parameters of our SRS-LBP.
All experiments in this section were performed on the ICHFR 2012 dataset.

\begin{figure}
\centerline{\includegraphics[width=.45\textwidth]{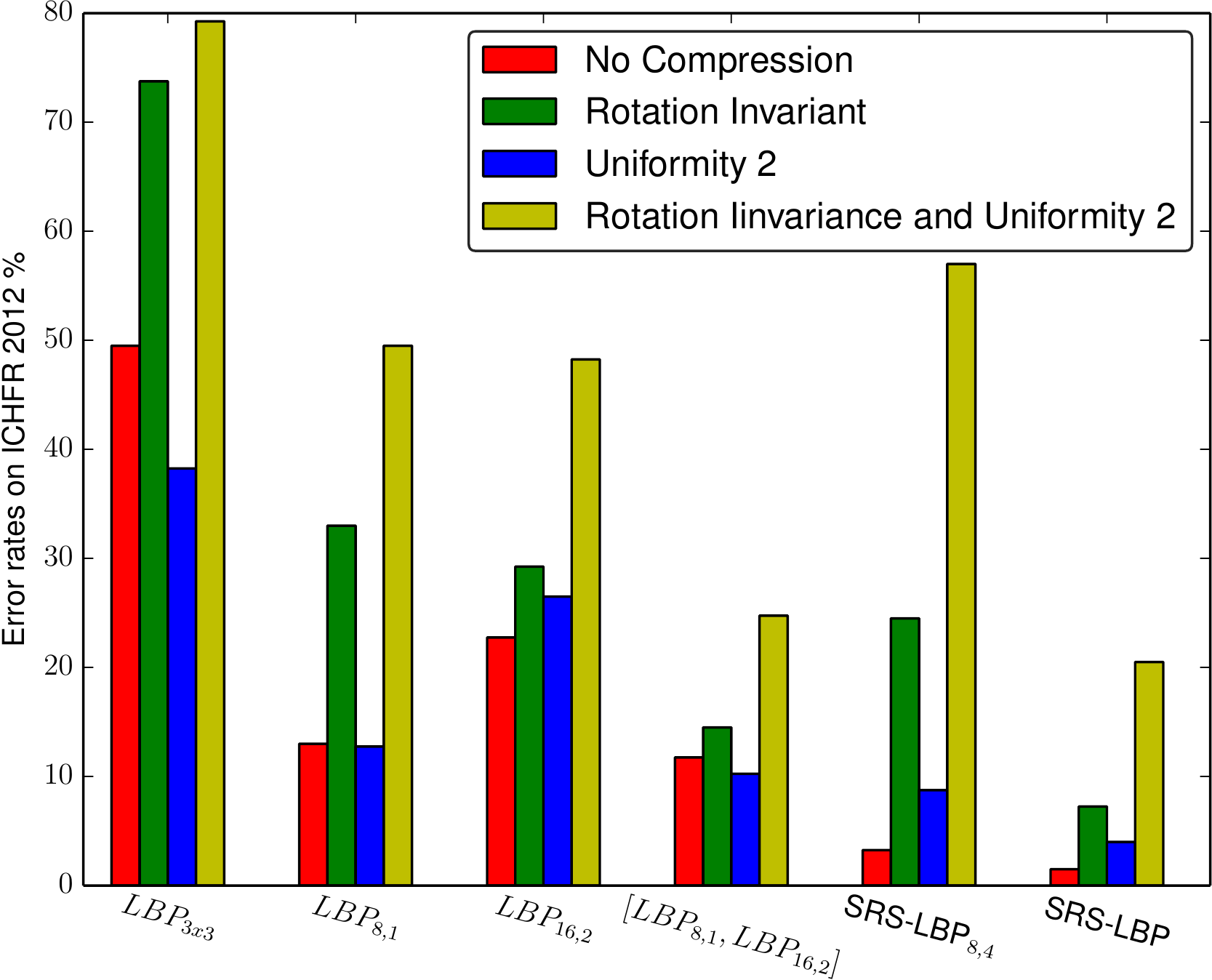}}
\vspace{-.2cm}
\caption{Comparison of SRS-LBP with baseline LBP.}
\label{fig:baseline}
\end{figure}
\minisection{Comparison of SRS-LBP and standard LBP.}  We consider
four standard LBP variants in these experiments:
$\mbox{LBP}_{3\times3}$~\cite{topi2000lbp},
$\mbox{LBP}_{8,1}$~\cite{ojala2002lbp},
$\mbox{LBP}_{16,2}$~\cite{ojala2002lbp}, and a concatenation of LBP
$[\mbox{LBP}_{8,1}, \mbox{LBP}_{16,2}]$~\cite{ojala2002lbp}. For
SRS-LBP we consider two variations: a single radius
$\mbox{SRS-LBP}_{8,4}$ and a multi-radius
$[\mbox{SRS-LBP}_{8,1},...,\mbox{SRS-LBP}_{8,12}]$.

In Fig.~\ref{fig:baseline} we report the 
error rates of top-1 accuracy leave-one-cross-validation for standard LBP and proposed SRS-LBP.
In addition to the full LBP vocabulary, we also show results for each LBP variant with combinations of vocabulary compression commonly employed for standard LBP. 
From this figure, we see that SRS-LBP outperform standard LBP in all compression modalities. 
Note also that SRS-LBP are less sensitive to compression than standard LBP.
Compression is usually applied to standard LBP in order to render vocabulary sizes tractable.
Due to their sparse nature, SRS-LBP are already a compact and tractable without resorting to compression.

\minisection{Radii Contribution.}
To better understand the contribution of each radius and principal component to the robustness of our descriptor, we decomposed performance as a function of each. 
In Fig.~\ref{fig:qualitative}(a) we show the accuracy of each radius independently and cumulatively (i.e. by concatenation). 
From this, we see that even large radii continue to contribute to improved recognition performance. 
Note also how, as radii grow beyond 3, the performance of the uniform-compressed features (in blue), drops quite sharply compared to the non-compressed features (in black).
This hints that uniformity compression doesn't scale to large radii. 
Since performance continues increasing until about twelve sparsely sampled radii, we use this configuration of SRS-LBP $=[\textrm{LBP}_{8,1},\ldots,\textrm{LBP}_{8,12}] $ for all subsequent experiments.

\minisection{Contribution of principal components.}  PCA in the
pipeline pipeline improves recognition accuracy in all cases.  In
Fig.~\ref{fig:qualitative}(b) we plot of top-1 writer identification
accuracy as a function of increasing PCs. Although performance quickly
begins to saturate, it improves steadily until around 200 principal
components. For all subsequent experiments we use 200 principal
components, resulting in a very compact image descriptor of only 200
dimensions.

\minisection{Rotation sensitivity.} Tolerance to small rotations can
be important for text documents, and in the case of handwritten text
the exact orientation of text is probably unknown. To measure
sensitivity to rotation, the ICHFR 2012 data-set was rotated from
angles -20\degree to 20\degree. For each rotation, all samples in the
non-rotated dataset were used as queries against the rotated dataset.
In Fig.~\ref{fig:qualitative}(c) we show the sensitivity to rotation
of the proposed SRS-LBP pipeline. An interesting observation in
Fig.~\ref{fig:qualitative}(b) is that while the single radius SRS-LBP
performs nearly equivalently to the multi-radius one, it is more
sensitive to rotations.

\subsection{Comparison with the State-Of-The-Art}

\begin{table}
\centering
\caption{Comparison with the SOA on ICDAR 2013}
\vspace{-.2cm}
\label{tbl:demokritos2013all}
\setlength{\tabcolsep}{3pt}
\begin{tabular}{|l||c|c|c|c||c|c|c|}
\hline
\textbf{Method} & \textbf{Top 1} & \textbf{Top 2} & \textbf{Top 5} & \textbf{Top 10} & \textbf{Hard 2} & \textbf{Hard 3} \\ \hline\hline
Tebessa-C*\hfill\cite{djeddi2012} & 93.4 & 96.1 & 97.8 & 98.0 & 62.6 & 37.8\\ \hline
CS-UMD-a*\hfill\cite{jain2013}   & 95.1 & 97.7 & 98.6 & 99.1& 19.6 & 7.1\\ \hline
Super Vector\hfill\cite{supervector2014}  & 97.1 & NA & NA & NA & 42.8 & 23.8 \\ \hline\hline
$\textrm{SRS-LBP}_{8,4}$ \textbf{l1out} &  97.2 & 98.2 & 98.9 & 99.2& 52.9 & 29.2 \\ \hline
SRS-LBP \textbf{metric}* & 96.9 & 98.5 & 99.0 & 99.5  & 54.5 & 32.9\\ \hline 
\end{tabular}
\end{table}

\begin{table}
\centering
\caption{State-of-the-art on ICDAR 2013 Greek}
\vspace{-.2cm}
\label{tbl:demokritos2013gr}
\begin{tabular}{|l||c|c|c|c|}
\hline
\textbf{Method} & \textbf{Top 1} & \textbf{Top 2} & \textbf{Top 5} & \textbf{Top 10} \\ \hline\hline
Tebessa-C*\hfill\cite{djeddi2012} & 93.1 & 97.0 & 99.5 & 99.5 \\ \hline
Delta-n Hinge\hfill\cite{delta2014} & 93.4 & NA & NA & 98.4 \\ \hline
CS-UMD-a*\hfill\cite{jain2013}  &  95.1 & 97.7 & 98.6 & 99.1\\ \hline 
Multi Feature*\hfill\cite{jain2014} & \textbf{99.2} & \textbf{99.6} & \textbf{99.8} & \textbf{99.8} \\ \hline\hline
$\textrm{SRS-LBP}_{8,4}$ \textbf{l1out} & 96.6 & 98.0 & 99.6 & 99.8 \\ \hline 
SRS-LBP \textbf{metric}* & 96.6 & 97.8 & 98.8 & 99.4\\ \hline 
SRS-LBP \textbf{l1out}& 98.4 & 99.2 & 99.4 &\textbf{99.8} \\ \hline 
\end{tabular}
\end{table}

\begin{table}
\centering
\caption{State-of-the-art on ICDAR 2013 English}
\label{tbl:demokritos2013en}
\vspace{-.2cm}
\begin{tabular}{|l||c|c|c|c|}
\hline
\textbf{Method} & \textbf{Top 1} & \textbf{Top 2} & \textbf{Top 5} & \textbf{Top 10} \\ \hline\hline
Tebessa-C*\hfill \cite{djeddi2012} &91.5 & 95.5 & 97.5 & 98.0 \\ \hline
Delta-n Hinge \hfill \cite{delta2014} & 93.4 & NA & NA & 97.8 \\ \hline
CS-UMD-a* \hfill \cite{jain2013}  &  95.2 & 98.2 & 98.8 & 99.2\\ \hline
Multi Feature* \hfill \cite{jain2014} & \textbf{97.4} & \textbf{97.8} & \textbf{98.6} & 98.8 \\ \hline\hline
$\textrm{SRS-LBP}_{8,4}$ \textbf{l1out} &95.2 & 96.4 & 98.0 & 98.4 \\ \hline
SRS-LBP \textbf{metric}* &  95.6 & 96.8 & 98.4 & \textbf{99.0}\\ \hline 
\end{tabular}
\end{table}

\begin{table*}
\centering
\caption{State-of-the-art on CVL using only 4+1 samples per writer}
\vspace{-.15cm}
\label{tbl:cvl}
\begin{tabular}{|l||c|c|c|c||c|c|c|}
\hline
\textbf{Method} & \textbf{Soft Top 1} & \textbf{Soft Top 2} & \textbf{Soft Top 5} & \textbf{Soft Top 10} & \textbf{Hard Top 2} & \textbf{Hard Top 3} & \textbf{Hard Top 4} \\ \hline\hline

Tebessa-C* \hfill \cite{djeddi2012} & 97.6 & 97.9 & 98.3 & 98.5  & 96.1 & 94.2 & 90.0 \\ \hline
Multi-feature* \hfill \cite{jain2014}& \textbf{99.4}& \textbf{99.5}& \textbf{99.6} &\textbf{99.7} &  98.3  & 94.8 & 82.9 \\ \hline
Super Vector \hfill \cite{supervector2014}  & 99.2 & NA & NA & NA  & 98.1 & 95.8 & 88.7 \\ \hline\hline
$\textrm{SRS-LBP}_{8,4}$ \textbf{l1out}& 99.0 & 99.2 & 99.4 & 99.5 &  97.7 & 95.2 & 86.0 \\ \hline 
SRS-LBP \textbf{metric}* & 98.6 & 98.8 & 98.9 & 99.1 & 97.8 & 94.6 & 85.3 \\ \hline 
SRS-LBP \textbf{l1out} & \textbf{ 99.4 } & 99.4 & 99.5 & 99.6 &  \textbf{98.6} & \textbf{97.0} & \textbf{90.1} \\ \hline 
\end{tabular}
\end{table*}

In recent years the topic of writer identification has seen a lot of
activity.  Contests, datasets, as well as several top performing
methods have been published. In this section we compare SRS-LBP with
the SOA in writer identification.

\minisection{Performance on ICDAR 2013.}  In
Table~\ref{tbl:demokritos2013all} we compare the performance of
SRS-LBP with the SOA on the ICDAR 2013 contest
dataset. In Table~\ref{tbl:demokritos2013gr} and
Table~\ref{tbl:demokritos2013en} we compare our performance with the
SOA on the Greek and English portions of the ICDAR 2013
dataset. As of this writing, the Multi Feature method represents the
SOA in writer identification~\cite{jain2014}. This
technique is based on character segmentation and clustering (which is
one reason they do not report results on the mixed-language dataset)
and multiple features extracted from characters. It is interesting
that our approach, which is based on dense extraction of a single
feature, performs comparably to this more complicated technique.

\minisection{Performance on CVL.}  Finally, in Table~\ref{tbl:cvl} we
compare the performance on SRS-LBP with the SOA on the
CVL dataset. On this dataset we have the most complete comparison with
SOA approaches for both hard and soft criteria. Our
approach performs equivalently to the Multi Feature technique
of~\cite{jain2014} for top-1 evaluation criterion, and we outperform
all others approaches for the hard recognition evaluation criterion
across all ranks which is associated with writer retrieval.

\section{Conclusion}
\label{sec:conclusion}

In this paper we introduced a writer identification approach based on sparse radial sampling Local Binary Patters. Our approach achieves SOA performance on ICDAR 2013 and CVL datasets. Our proposed method has several advantages over other SOA techniques. Of the top performers, ours is the only one that is based on dense extraction of a single local feature descriptor. This makes it applicable at the earliest stages in a DIA pipeline without the need for segmentation, binarization, or extraction of multiple features. Its simplicity makes it a good candidate for fusion with other methods.

The SRS-LBP we propose is a generic LBP variant that should be applicable to many document image analysis tasks such as font recognition, handwritten and printed script detection, ant document page classification. Future and ongoing work will concentrate on experiments demonstrating the range of application the proposed method has and the improvements it can bring in end-to-end document recognition scenarios.  While PCA was the sole learning component of the proposed method, it is unsupervised and feature space transformations taking full advantage of labelled training data should yield improved results.

\newcommand{\BIBdecl}{\setlength{\itemsep}{0.10 em}}
\bibliographystyle{IEEEtran}
\bibliography{writerid}

\end{document}